\documentclass[a4paper,fleqn]{cas-dc}

\usepackage[numbers]{natbib}

\usepackage{booktabs}
\usepackage{multirow}
\usepackage{graphicx}
\usepackage{caption}
\usepackage{subcaption}

\def\tsc#1{\csdef{#1}{\textsc{\lowercase{#1}}\xspace}}
\tsc{WGM}
\tsc{QE}
\tsc{EP}
\tsc{PMS}
\tsc{BEC}
\tsc{DE}

\begin{document}
\let\WriteBookmarks\relax
\def\floatpagepagefraction{1}
\def\textpagefraction{.001}

\shorttitle{SSL for Knee OA}
\shortauthors{Haresh Rajamohan et~al.}

\title [mode = title]{Self-Supervised Learning for Knee Osteoarthritis: Diagnostic Limitations and Prognostic Value of Hospital Data}


\author[1]{Haresh Rengaraj Rajamohan}
\cormark[1] 
\ead{hrr288@nyu.edu} 
\credit{Conceptualization, Methodology, Software}

\author[4]{Yuxuan Chen}
\ead{Yuxuan.Chen@nyulangone.org}
\credit{Data curation, Validation}

\author[1]{Kyunghyun Cho}
\ead{kc119@nyu.edu}
\credit{Supervision, Methodology}

\author[3,2]{Cem M. Deniz}
\ead{cem.deniz@nyulangone.org}
\credit{Supervision, Funding acquisition}


\affiliation[1]{organization={Center for Data Science, New York University},
                city={New York},
                postcode={10011},
                state={NY},
                country={United States}}

\affiliation[2]{organization={Bernard and Irene Schwartz Center for Biomedical Imaging, New York University Langone Health},
                city={New York},
                postcode={10016}, 
                state={NY},
                country={USA}}

\affiliation[3]{organization={Department of Radiology, New York University Langone Health},
                city={New York},
                postcode={10016},
                state={NY},
                country={USA}}
                
\affiliation[4]{organization={Perlmutter Cancer Center, New York University Langone Health},
                city={New York},
                postcode={10016},
                state={NY},
                country={USA}}

\cortext[cor1]{Corresponding author}


\begin{abstract}
This study assesses whether self-supervised learning (SSL) improves knee osteoarthritis (OA) modeling for diagnosis and prognosis relative to ImageNet-pretrained initialization. We compared (i) image-only SSL pretrained on knee radiographs from the OAI, MOST, and NYU cohorts, and (ii) multimodal image-text SSL pretrained on hospital knee radiographs paired with radiologist impressions. For diagnostic Kellgren-Lawrence (KL) grade prediction, SSL yielded mixed results. While image-only SSL improved accuracy during linear probing (frozen encoder), it did not outperform ImageNet pretraining during full fine-tuning. Similarly, multimodal SSL failed to improve grading performance. A likely explanation is mismatch between the hospital pretraining corpus and the downstream diagnostic task: the hospital image-text dataset was restricted to knees from patients with clinically identified OA in routine care, rather than a cohort spanning the full spectrum from normal to severe disease needed for balanced KL grading. In addition, radiology impressions do not explicitly encode KL grade, limiting supervision for learning KL-specific decision boundaries. In contrast, this same multimodal initialization significantly improved prognostic modeling. It outperformed ImageNet baselines in predicting 4-year structural incidence and progression, including on external validation (MOST AUROC: 0.701 vs. 0.599 at 10\% labeled data). Overall, these results suggest that our hospital image-text data may be less effective for diagnostic grading when the pretraining cohort is limited to OA knees, but can provide a strong signal for prognostic modeling when the downstream task is better aligned with the pretraining data distribution.
\end{abstract}

\begin{highlights}
\item Image-only self-supervised learning improved linear-probe performance (0.422 vs.\ 0.320 Acc) but did not outperform ImageNet pretraining for knee OA diagnosis under full fine-tuning (0.636 vs.\ 0.646 Acc).
\item Multimodal pretraining on hospital radiographs paired with radiology impressions did not improve KL grading performance (ConVIRT 0.627 vs.\ ImageNet 0.646 Acc), suggesting limited alignment between the hospital pretraining cohort and the downstream diagnostic task.
\item The same multimodal initialization improved prognostic modeling, outperforming ImageNet on structural progression prediction, including on external validation (AUROC 0.701 vs.\ 0.599).
\end{highlights}


\maketitle

\section{Introduction}

Knee Osteoarthritis (OA) is a progressive, debilitating joint disorder and a leading cause of disability worldwide. Its gradual onset and potential for irreversible structural damage make early intervention crucial for optimizing patient outcomes and reducing long-term healthcare costs \cite{cross2014global}. However, the clinical management of knee OA faces a critical bottleneck: accurately predicting prognosis. Specifically, identifying which patients will experience rapid structural deterioration remains a significant challenge. Reliable prognostic models are essential for guiding timely behavioral interventions and optimizing patient selection for clinical trials, yet their development is hindered by the scarcity of labeled longitudinal data.

Deep learning (DL) has fundamentally transformed medical image analysis, demonstrating diagnostic performance comparable to human experts across domains ranging from diabetic retinopathy detection to dermatological lesion classification \cite{gulshan2016development,ting2017development,irvin2019chexpert,esteva2017dermatologist}. DL has also shown immense promise for automating knee OA assessment. Foundational works by \citet{tiulpin2018automatic,zhang2020attention} successfully leveraged convolutional neural networks (CNNs) to automate disease severity grading (Kellgren-Lawrence [KL] grading \cite{kellgren1957radiological}) from plain radiographs. Building on this diagnostic success, recent research has pivoted toward complex prognostic tasks, such as predicting total knee replacement (TKR) risk or forecasting disease progression using knee radiographs and MRIs \cite{tiulpin2019multimodal,tolpadi2020deep,rajamohan2023prediction,leung2020prediction,rajamohan2025progressive}. However, a fundamental disparity exists in data availability. While diagnostic datasets can be assembled from independent one-time imaging exams across patients, prognostic datasets require years of continuous patient follow-up to capture disease trajectories.

To address label scarcity, self-supervised learning (SSL) has emerged as a transformative paradigm in deep learning. By defining pretext tasks, such as instance discrimination or image-text matching, SSL allows models to learn robust representations from vast repositories of unlabeled data. In the natural image domain, contrastive methods (e.g., SimCLR, MoCo) and non-contrastive methods (e.g., Barlow Twins, ViCReg, DINO) have demonstrated that SSL pretraining can rival or exceed supervised pretraining on ImageNet classification, particularly in low-labeled data regimes \cite{chen2020simple,chen2020improved,zbontar2021barlow,bardes2021vicreg,caron2021emerging}. Furthermore, multimodal approaches like CLIP \cite{radford2021learning}, which leverage image-caption pairs, have pushed the boundaries of zero-shot transfer and downstream classification performance.

These advancements are particularly relevant in the medical domain, where hospitals generate vast volumes of unlabeled scans. Pioneering works have adapted SSL to tasks ranging from chest X-ray interpretation to dermatology, often demonstrating that ``in-domain'' pretraining yields superior downstream performance compared to standard ImageNet \cite{deng2009imagenet} initialization \cite{sowrirajan2021moco,azizi2021big,zhang2022contrastive,huang2021gloria,huang2024radiology}. A key open question is whether in-domain SSL using routine-care knee radiographs (with or without paired reports) provides reliable gains over ImageNet for both cross-sectional severity grading and longitudinal prognosis.

However, the application of SSL to knee OA faces a practical dilemma regarding data availability and distribution. In routine-care hospital data, standardized KL grades are typically not recorded, and even when proxy KL labels can be obtained, the radiograph distribution is strongly spectrum-biased because imaging is ordered for symptomatic patients. As a result, learning a general-purpose KL grader requires curated cohorts with broad severity coverage (e.g., OAI and MOST), which are expensive to collect and limited in scale. In contrast, radiologist impressions are produced as part of standard clinical workflow and are available at hospital scale. This accessibility motivates multimodal SSL as a practical alternative for representation learning, despite the inherent spectrum bias of the underlying data.

In this study, we analyze when SSL pretraining improves knee OA modeling under realistic limited labeled data fine-tuning protocols. First, we show that standard image-only SSL strategies, even when trained on diverse in-domain knee radiographs (OAI and MOST datasets), do not consistently outperform ImageNet initialization for KL grading under fine-tuning, despite improved linear probing. Notably, this limitation was not unique to knee OA; we observed similar stagnation when replicating these image-only experiments on chest x-rays from NIH chest radiograph dataset. Second, we evaluate multimodal contrastive pretraining (ConVIRT, GLoRIA) on routine-care hospital knee radiographs paired with impression text. While ConVIRT multimodal pretraining reproduces expected gains in chest radiography on CheXpert, it does not improve KL grading for knee OA. One likely reason is task-data mismatch: our hospital knee corpus was restricted to patients with clinically identified knee osteoarthritis who underwent imaging in routine care, so we only had access to OA knees rather than the full spectrum from normal to severe disease needed for balanced KL grading. In addition, impression text does not explicitly encode KL grade, limiting supervision for learning KL decision boundaries.

Finally, we test whether multimodal pretraining is better matched to prognostic endpoints that require discrimination among similarly diseased knees. We evaluate 4-year structural incidence and progression prediction on OAI and external validation on MOST, comparing multimodal SSL against ImageNet and an upper-bound supervised initialization pretrained on expert KL labels from OAI. We find that multimodal SSL significantly improves prognosis performance across all labeled-data fractions and enhances discrimination of progressors within each KL strata, indicating robust performance improvements. Together, these findings highlight that routine-care image-text data may be poorly suited for learning general OA severity graders, yet can provide a scalable and effective signal for prognostic modeling when aligned with the downstream task.

\section{Materials and Methods}

\subsection{Data Sources}
This retrospective study utilized data from three independent sources: the Osteoarthritis Initiative (OAI), the Multicenter Osteoarthritis Study (MOST), and a dataset from NYU Langone Health.

The \textbf{Osteoarthritis Initiative (OAI)} is a multicenter, prospective observational study of knee OA \cite{nevitt2006osteoarthritis}. We obtained bilateral knee radiographs and clinical data spanning baseline to 108-month follow-up. The OAI dataset served as the primary source for downstream evaluation for diagnosis and prognosis tasks.

The \textbf{Multicenter Osteoarthritis Study (MOST)} is a longitudinal observational study of individuals with or at high risk of knee OA \cite{segal2013multicenter} evaluated at baseline to 168-month follow-up. MOST data was utilized as an external validation cohort to assess the generalizability of the diagnosis and prognosis models.

\textbf{NYU Langone Health Dataset:}
This dataset represents routine clinical data acquired from a tertiary care center. Retrospective data was collected from the electronic medical records and PACS of NYU Langone Health spanning January 1, 2011, to September 1, 2017 (IRB No. i17-01339). The cohort was defined using adult patients with knee osteoarthritis diagnosis codes who underwent knee radiography or MRI as part of routine clinical care. Patients with incomplete demographic data (age, sex, BMI, ethnicity) were excluded. This dataset was fully de-identified and utilized exclusively for self-supervised pretraining.

For the chest experiments, we utilized MIMIC-CXR \cite{johnson2019mimic}, CheXpert \cite{irvin2019chexpert}, and NIH Chest X-ray 14 \cite{wang2017chestx} datasets. The multimodal pretraining was performed using image-text pairs from MIMIC-CXR and downstream task evaluation was performed on CheXpert. For the image only self-supervised training, NIH Chest X-ray 14 was used for both pretraining and downstream evaluation.

\subsection{Data Partitioning and Preprocessing}

\subsubsection{Pretraining Datasets}
To evaluate the impact of data distribution, we curated two distinct pretraining datasets:
\begin{itemize}
    \item \textbf{Image-Only Pretraining Dataset ($D_{SSL-Image}$):} A large-scale, diverse dataset comprising 57,325 extracted bilateral knee joint radiographs from OAI, MOST, and NYU. To prevent data leakage, patients assigned to the validation or test sets of the diagnosis and prognosis downstream tasks (OAI/MOST) were strictly excluded. 
    \item \textbf{Multimodal Pretraining Dataset ($D_{SSL-Multi}$):} A dataset comprising 84,294 pairs of knee radiographs and free-text radiology impressions exclusively from NYU Langone Health. Unlike $D_{SSL-Image}$, this dataset reflects the natural selection bias of a tertiary care center. The acquisition protocols were heavily skewed toward osteoarthritis assessment, with the vast majority of exams consisting of weight-bearing anteroposterior (AP) or posteroanterior (PA) flexion views, accompanied by lateral and patellar (sunrise) projections. Bilateral studies dominated the cohort, though unilateral and portable bedside acquisitions were also present. Analysis of the disease distribution in bilateral knee joints (inferred via a supervised proxy model) estimated that approximately 93\% of this cohort represents KL Grade 3.
\end{itemize}

\subsubsection{Image and Text Preprocessing}
For all datasets, knee joints were localized and cropped from bilateral radiographs. We utilized a ResNet-based localization model trained to directly regress bounding box coordinates for the knee joints \cite{zhang2020attention}. Based on these predictions, regions of interest (ROIs) of size $1024 \times 1024$ pixels were extracted. These ROIs were subsequently resized to $256 \times 256$ pixels, followed by random cropping to $224 \times 224$ pixels and normalization during training.
For multimodal learning, text impressions were extracted from radiology reports. We deidentified and removed PHI from the text using \cite{datta2020new}, applied standard text cleaning (lowercasing, removal of special characters) and tokenization using the WordPiece tokenizer with a maximum sequence length of 512 tokens.

\subsection{Model Development and Pretraining}
We utilized a ResNet34 \cite{he2016deep} backbone for all the knee experiments similar to previous works \cite{zhang2020attention} (ResNet50 for the Chest experiments). We implemented two categories of self-supervised learning (SSL):

\subsubsection{Image-Only SSL}
We trained models on $D_{SSL-Image}$ using four approaches:
\begin{itemize}
    \item \textbf{MoCo:} Utilized a momentum encoder and a queue of negative samples to maximize similarity between augmented views of the same image \cite{chen2020improved}.
    \item \textbf{Barlow Twins:} Minimized the cross-correlation matrix difference between distorted views to identity \cite{zbontar2021barlow} without requiring negative pairs.
    \item \textbf{ViCReg:} Enforced variance, invariance, and covariance regularization without requiring negative pairs \cite{bardes2021vicreg}.
    \item \textbf{CNN-JEPA:} A non-contrastive, predictive SSL objective that learns representations by predicting latent embeddings of masked/held-out regions from context, encouraging invariant and information-rich features without explicit negative pairs \cite{kalapos2024cnn,assran2023self}.
\end{itemize}

\subsubsection{Multimodal SSL}
We trained models on $D_{SSL-Multi}$ using image-text pairs:
\begin{itemize}
    \item \textbf{ConVIRT:} Optimized a bidirectional contrastive loss ($L_{InfoNCE}$) to maximize the similarity between the image embedding and its corresponding text embedding relative to other pairs in the batch \cite{zhang2022contrastive}.
    \item \textbf{GLORIA:} Utilized a global-local attention mechanism to align specific image sub-regions with word-level text features, hypothesized to capture localized pathology \cite{huang2021gloria}.
\end{itemize}

\subsection{Downstream Evaluation Tasks}

\subsubsection{Knee Osteoarthritis (OA)}
We evaluated the models on two primary clinical tasks using the OAI and MOST datasets:

\begin{itemize}
    \item \textbf{Diagnosis:} Classification based on Kellgren-Lawrence Grade (KLG) \cite{kellgren1957radiological} -- A severity scale (0-4) assigned by radiologists based on radiographic features including joint space narrowing, bone spurs, and sclerosis.
    
    \item \textbf{Prognosis:} Prediction of disease worsening over a 4-year horizon \cite{lee2018prevalence}, defined as:
    \begin{itemize}
        \item \textit{Structural Incidence} for early-stage OA (KLG 0-1): Progression to KLG $\ge$ 2 or undergoing a Total Knee Replacement (TKR).
        \item \textit{Structural Progression} for radiographic OA (KLG $\ge$ 2): An increase in KLG or undergoing a TKR.
    \end{itemize}
\end{itemize}

\begin{table*}[t]
\centering
\caption{\textbf{Patient Characteristics in Structural Progression Cohorts (OAI and MOST).} Data are presented as mean $\pm$ standard deviation where applicable. BMI = Body Mass Index.}
\label{tab:progression_demographics}
\resizebox{0.7\textwidth}{!}{%
\begin{tabular}{llcccc}
\toprule
& & \multicolumn{2}{c}{\textbf{Men}} & \multicolumn{2}{c}{\textbf{Women}} \\
\cmidrule(lr){3-4} \cmidrule(lr){5-6}
\textbf{Dataset} & \textbf{Parameters} & \textbf{Patients} & \textbf{Controls} & \textbf{Patients} & \textbf{Controls} \\
\midrule
\multirow{10}{*}{\textbf{OAI}} 
 & No. of patients & 216 & 707 & 355 & 970 \\
 & No. of scans & 241 & 917 & 421 & 1348 \\
 & Mean age (y) & 62.3 $\pm$ 9.0 & 62.1 $\pm$ 9.3 & 63.7 $\pm$ 8.2 & 62.5 $\pm$ 8.9 \\
 & Mean height (m) & 1.8 $\pm$ 0.1 & 1.8 $\pm$ 0.1 & 1.6 $\pm$ 0.1 & 1.6 $\pm$ 0.1 \\
 & Mean weight (kg) & 93.1 $\pm$ 14.3 & 91.7 $\pm$ 14.4 & 80.8 $\pm$ 14.8 & 78.2 $\pm$ 14.5 \\
 & Mean BMI (kg/m$^2$) & 30.1 $\pm$ 4.0 & 29.3 $\pm$ 4.0 & 30.6 $\pm$ 5.2 & 29.6 $\pm$ 5.3 \\
 \cmidrule(l){2-6}
 & \textit{Ethnicity} & & & & \\
 & \hspace{3mm} White & 184 & 595 & 253 & 721 \\
 & \hspace{3mm} Black & 26 & 97 & 90 & 230 \\
 & \hspace{3mm} Asian & 2 & 4 & 3 & 5 \\
 & \hspace{3mm} Other nonwhite & 4 & 11 & 9 & 14 \\
\midrule
\multirow{9}{*}{\textbf{MOST}} 
 & No. of patients & 203 & 231 & 359 & 386 \\
 & No. of scans & 231 & 282 & 427 & 495 \\
 & Mean age (y) & 62.8 $\pm$ 8.5 & 63.2 $\pm$ 8.4 & 64.2 $\pm$ 7.6 & 63.5 $\pm$ 7.3 \\
 & Mean height (m) & 1.8 $\pm$ 0.1 & 1.8 $\pm$ 0.1 & 1.6 $\pm$ 0.1 & 1.6 $\pm$ 0.1 \\
 & Mean weight (kg) & 100.9 $\pm$ 19.8 & 98.3 $\pm$ 17.8 & 86.3 $\pm$ 18.5 & 85.0 $\pm$ 18.2 \\
 & Mean BMI (kg/m$^2$) & 32.0 $\pm$ 5.8 & 31.1 $\pm$ 5.4 & 32.7 $\pm$ 6.8 & 31.9 $\pm$ 6.6 \\
 \cmidrule(l){2-6}
 & \textit{Ethnicity} & & & & \\
 & \hspace{3mm} White & 177 & 197 & 288 & 302 \\
 & \hspace{3mm} Black & 22 & 32 & 63 & 82 \\
 & \hspace{3mm} Other & 4 & 2 & 8 & 2 \\
\bottomrule
\end{tabular}%
}
\end{table*}

\begin{table*}[t]
\centering
\caption{\textbf{Patient Characteristics in Structural Incidence Cohorts (OAI and MOST).} Data are presented as mean $\pm$ standard deviation where applicable.}
\label{tab:incidence_demographics}
\resizebox{0.7\textwidth}{!}{%
\begin{tabular}{llcccc}
\toprule
& & \multicolumn{2}{c}{\textbf{Men}} & \multicolumn{2}{c}{\textbf{Women}} \\
\cmidrule(lr){3-4} \cmidrule(lr){5-6}
\textbf{Dataset} & \textbf{Parameters} & \textbf{Patients} & \textbf{Controls} & \textbf{Patients} & \textbf{Controls} \\
\midrule
\multirow{10}{*}{\textbf{OAI}} 
 & No. of patients & 149 & 1103 & 271 & 1336 \\
 & No. of scans & 160 & 1737 & 296 & 2171 \\
 & Mean age (y) & 61.0 $\pm$ 8.7 & 59.6 $\pm$ 9.4 & 60.3 $\pm$ 8.6 & 60.5 $\pm$ 9.0 \\
 & Mean height (m) & 1.8 $\pm$ 0.1 & 1.8 $\pm$ 0.1 & 1.6 $\pm$ 0.1 & 1.6 $\pm$ 0.1 \\
 & Mean weight (kg) & 91.0 $\pm$ 14.9 & 91.0 $\pm$ 14.9 & 77.2 $\pm$ 14.2 & 71.3 $\pm$ 13.3 \\
 & Mean BMI (kg/m$^2$) & 29.1 $\pm$ 4.1 & 28.1 $\pm$ 3.8 & 29.3 $\pm$ 4.8 & 27.1 $\pm$ 4.8 \\
 \cmidrule(l){2-6}
 & \textit{Ethnicity} & & & & \\
 & \hspace{3mm} White & 133 & 975 & 208 & 1121 \\
 & \hspace{3mm} Black & 13 & 110 & 56 & 180 \\
 & \hspace{3mm} Asian & 1 & 4 & 4 & 13 \\
 & \hspace{3mm} Other nonwhite & 2 & 12 & 2 & 21 \\
\midrule
\multirow{9}{*}{\textbf{MOST}} 
 & No. of patients & 179 & 600 & 326 & 795 \\
 & No. of scans & 196 & 945 & 371 & 1304 \\
 & Mean age (y) & 61.3 $\pm$ 7.8 & 60.6 $\pm$ 7.8 & 62.4 $\pm$ 8.0 & 61.1 $\pm$ 7.7 \\
 & Mean height (m) & 1.8 $\pm$ 0.1 & 1.8 $\pm$ 0.1 & 1.6 $\pm$ 0.1 & 1.6 $\pm$ 0.1 \\
 & Mean weight (kg) & 99.8 $\pm$ 17.1 & 93.7 $\pm$ 15.1 & 82.6 $\pm$ 15.6 & 77.4 $\pm$ 14.1 \\
 & Mean BMI (kg/m$^2$) & 31.3 $\pm$ 5.3 & 29.6 $\pm$ 4.5 & 30.9 $\pm$ 5.9 & 28.8 $\pm$ 5.2 \\
 \cmidrule(l){2-6}
 & \textit{Ethnicity} & & & & \\
 & \hspace{3mm} White & 157 & 573 & 270 & 694 \\
 & \hspace{3mm} Black & 19 & 73 & 50 & 89 \\
 & \hspace{3mm} Other & 3 & 10 & 3 & 12 \\
\bottomrule
\end{tabular}%
}
\end{table*}


    

\subsubsection{Chest Radiography (Control Experiments)}
To validate the methodology in a domain where SSL is established, we conducted two independent control experiments with distinct downstream evaluation tasks:

\begin{itemize}
    \item \textbf{Image-Only SSL:} We pretrained and fine-tuned image-only encoders (MoCo, ViCReg) using the \textbf{NIH Chest X-ray 14} dataset \cite{wang2017chestx}. The downstream task was multi-label classification of the 14 standard thoracic pathologies (e.g., Pneumothorax, Effusion) defined in the dataset. This assessed whether pixel-only pretraining offers gains when the pretraining and downstream data distributions are identical.
    
    \item \textbf{Multimodal SSL:} We pretrained multimodal encoders (ConVIRT) using the image-text pairs from the \textbf{MIMIC-CXR} dataset \cite{johnson2019mimic}. These models were then fine-tuned on the \textbf{CheXpert} dataset \cite{irvin2019chexpert} for the standard 5-disease classification task (Atelectasis, Cardiomegaly, Consolidation, Edema, and Pleural Effusion). This served to validate our multimodal implementation against established benchmarks.
\end{itemize}
\subsection{Experimental Pipeline and Evaluation Protocol}
\begin{figure}[h!]
\centering
\includegraphics[width=\columnwidth]{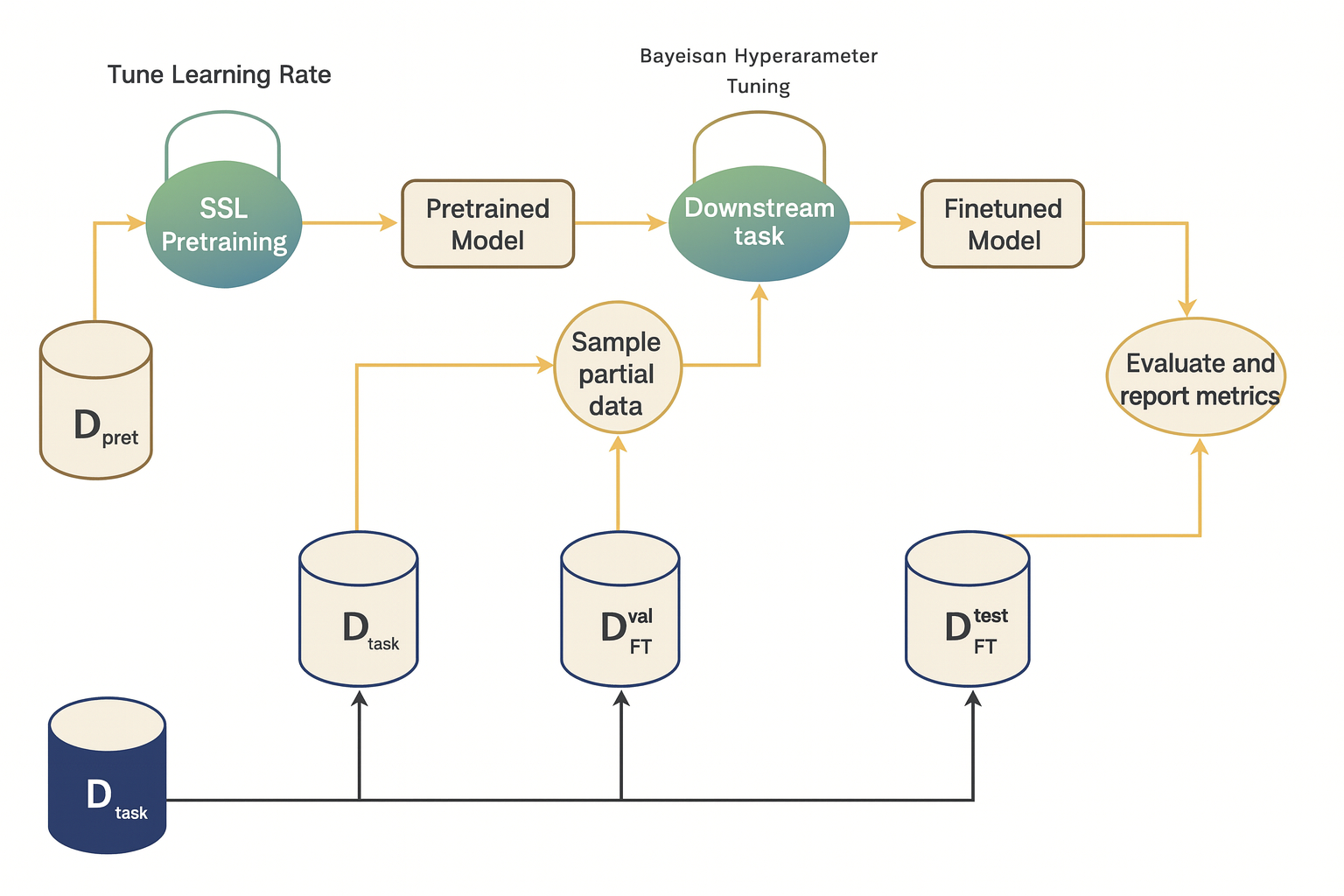} 
\caption{\textbf{Overview of the Experimental Pipeline.} The workflow consists of SSL pretraining followed by downstream fine-tuning. A critical feature of our protocol is the "Sample partial data" step: when evaluating low-data regimes (e.g., 1\% training data), we proportionally subsample \textit{both} the training and validation sets for hyperparameter tuning. This ensures a realistic evaluation, contrasting with prior works that utilize full validation sets even when training data is scarce.}
\label{fig:pipeline}
\end{figure}

The overall experimental workflow employed in this study is illustrated in Figure \ref{fig:pipeline}. The process is divided into two primary stages: (1) Self-supervised pretraining on unlabeled data ($D_{pret}$), and (2) Supervised fine-tuning on downstream task data ($D_{task}$).

\subsubsection{Realistic Low-Data Evaluation}
A critical aspect of our methodology is the rigorous handling of low-data regimes during fine-tuning. As shown in the "Sample partial data" block of Figure \ref{fig:pipeline}, when evaluating performance at specific data fractions (e.g., 1\%, 5\%, 10\%), we sample that fraction proportionally from \textbf{both} the available training ($D_{task}$) and validation ($D_{FT}^{val}$) pools.

This approach contrasts with common practices in prior medical SSL literature (e.g., some chest X-ray studies) \cite{sowrirajan2021moco,zhang2022contrastive}, where models trained on tiny fractions of data are often tuned using larger, fixed-size validation sets. We argue that such protocols are unrealistic; in real-world scenarios where training labels are scarce, labeled validation data is equally scarce. By constraining the validation set size proportionally, our protocol provides a more stringent and realistic assessment of a model's true data efficiency.

\subsection{Control Experiments: Investigating Image-Only Limitations}
To rigorously investigate why image-only SSL might fail in the medical domain, we conducted specific control studies disentangling dataset size and domain complexity.

\subsubsection{Effect of Pretraining Dataset Size}
We hypothesized that the performance gap between Medical SSL and ImageNet transfer might stem from the sheer size of pretraining dataset (ImageNet - 1.2M images). To test this, we conducted two scaling experiments:
\begin{itemize}
    \item \textbf{Natural Image Scaling:} We created random subsets of ImageNet matching the scale of our medical datasets (6k, 10k, 25k, 40k, 58k images). We pretrained MoCo encoders on these subsets and evaluated their downstream performance on KLG diagnosis.
    \item \textbf{In-Domain Scaling:} We pretrained encoders on random subsets of our Knee dataset (6k to 58k) to see if a larger in-domain dataset scales performance.
\end{itemize}
A positive relationship between pretraining scale and downstream performance would suggest that the current knee pretraining dataset is of insufficient size. Conversely, a performance plateau would indicate that the limiting factor is the intrinsic information content of the data or downstream task complexity, rather than dataset size.
\subsubsection{Effect of Domain Complexity}
To test if the performance limitations of SSL lies in the medical images themselves (e.g., subtle grayscale texture vs. distinct natural objects in focus), we utilized \textbf{Imagenette}, a subset of 10 easily discriminable ImageNet classes. We pretrained encoders using SSL on progressively larger subsets of the ImageNet training set and evaluated them on Imagenette classification. This object-centric control isolates the effect of natural-image pretraining scale under a simpler visual recognition task, and serves as a sanity check that our SSL training and evaluation pipeline behaves as expected before applying the same methodology to knee radiographs.
\subsection{Evaluation Setup}
For the downstream tasks, we evaluated performance using both \textbf{Linear Probing (LP)} (frozen backbone) and \textbf{Full Fine-Tuning (FT)}. As illustrated in Figure \ref{fig:pipeline}, to ensure robust comparisons:
\begin{itemize}
    \item \textbf{Bayesian Hyperparameter Tuning:} We utilized the \textbf{Optuna} framework \cite{akiba2019optuna} with Tree-structured Parzen Estimators (TPE) to optimize hyperparameters like learning rate, weight decay, optimizer and scheduler for every method-dataset-fraction combination on the proportional validation sets.
    \item \textbf{Model Selection and Seeds:} To ensure statistical robustness, experiments were repeated across 10 independent random seeds. Crucially, the random subsampling of training and validation sets was distinct for each seed, ensuring that reported metrics reflect performance stability across different data distributions. For each run, the final model checkpoint was selected based on the optimal metric achieved on the validation set.
    \item \textbf{Metrics:} Performance is reported as Balanced Accuracy for KLG diagnosis. For Knee OA prognosis, performance is reported as the standard Area Under the ROC Curve (AUROC). For the multi-label Chest radiography tasks, we report the Macro-averaged AUROC. All results represent the mean and standard deviation across multiple independent random seeds evaluated on the held-out test set ($D_{FT}^{test}$).
\end{itemize}

\section{Results}

\subsection{Image-Only SSL Fails to Outperform ImageNet pretrained Baseline}
We first investigated whether large-scale in-domain pretraining on radiographs alone ($D_{SSL-Image}$) could outperform standard transfer learning. Table \ref{tab:knee_image_only} summarizes the performance of four image-only SSL methods (MoCo, Barlow Twins, ViCReg, and CNN-JEPA) compared to ImageNet initialization and random initialization on the downstream task of KLG diagnosis.

\textbf{Lack of Advantage over ImageNet:} Both ImageNet and in-domain SSL substantially outperform random initialization, confirming that pretraining is beneficial in the low-label regime. However, contrary to the “in-domain is better” hypothesis, ImageNet remains a consistently strong baseline: under full fine-tuning (FT) at 10\% labeled data, ImageNet achieves 0.646, marginally outperforming the best in-domain method (MoCo: 0.636).

\textbf{Linear Probing vs. Fine-Tuning:} While in-domain SSL yielded higher Linear Probing (LP) accuracy than ImageNet (e.g., MoCo LP: 0.422 vs. ImageNet LP: 0.320 at 10\% data) consistently, this advantage disappeared with fine-tuning. 

\noindent We observe the same overall pattern under additional image-only controls (mixed-domain diversity and MAE pretraining; Appendix~\ref{appendix:A}--\ref{appendix:B}).

\begin{table*}[t]
\centering
\footnotesize
\caption{\textbf{Image-Only SSL vs. ImageNet for Knee OA Diagnosis.} KLG prediction performance (Balanced Accuracy) on OAI test set. Note that ImageNet pretraining (Imgnet-pret) consistently rivals or beats in-domain SSL under fine-tuning.}
\label{tab:knee_image_only}
\begin{tabular*}{\textwidth}{@{\extracolsep{\fill}}lcccccc}
\toprule
& \multicolumn{2}{c}{\textbf{1\% Data}} & \multicolumn{2}{c}{\textbf{5\% Data}} & \multicolumn{2}{c}{\textbf{10\% Data}} \\
\cmidrule(lr){2-3} \cmidrule(lr){4-5} \cmidrule(lr){6-7}
\textbf{Pretraining Approach} & \textbf{FT} & \textbf{LP} & \textbf{FT} & \textbf{LP} & \textbf{FT} & \textbf{LP} \\
\midrule
Random & 0.397 (0.042) & 0.200 & 0.555 (0.011) & 0.200 & 0.619 (0.015) & 0.200 \\
Imgnet-pret & \textbf{0.483 (0.040)} & 0.263 & \textbf{0.613 (0.014)} & 0.308 & \textbf{0.646 (0.009)} & 0.320 \\
MoCo (Knee) & 0.475 (0.050) & 0.368 & 0.606 (0.010) & 0.411 & 0.636 (0.012) & 0.422 \\
Barlow (Knee) & 0.425 (0.063) & 0.324 & 0.589 (0.018) & 0.355 & 0.620 (0.010) & 0.358 \\
ViCReg (Knee) & 0.471 (0.048) & 0.336 & 0.585 (0.018) & 0.374 & 0.635 (0.007) & 0.373 \\
CNN-JEPA (Knee) & 0.476 (0.043) & 0.253  & 0.606 (0.015) & 0.33 & 0.631 (0.009) & 0.375  \\

\bottomrule
\end{tabular*}
\end{table*}

\subsection{Analysis: Data scale and task complexity}

\subsubsection{The Data Scale Hypothesis}
We first tested if the performance limitations stemmed from insufficient pretraining data by analyzing the scaling laws for both natural and in-domain images.

\begin{figure}[t]
    \centering
    \includegraphics[width=\columnwidth]{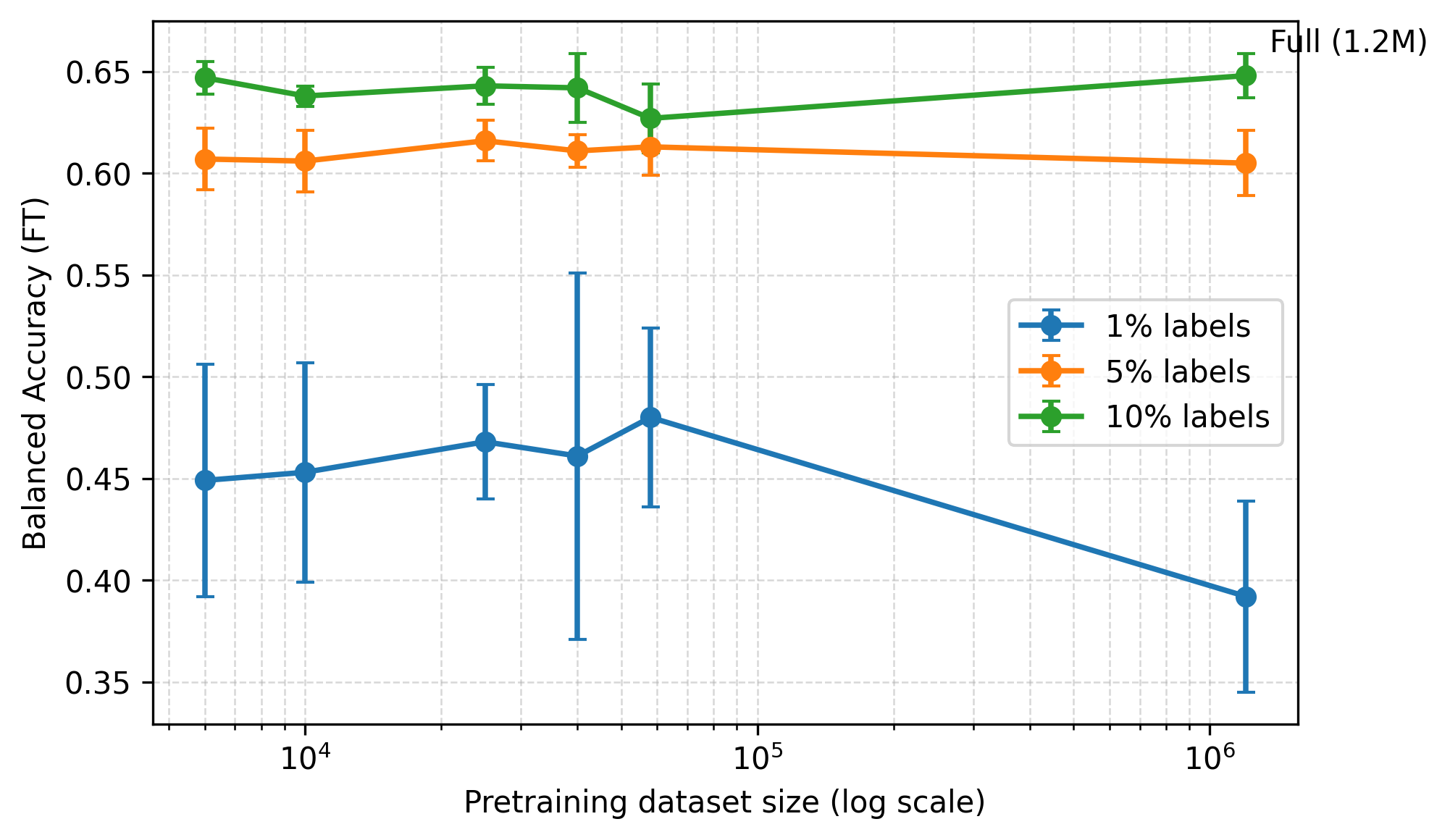}
    \caption{\textbf{ImageNet scaling.} Downstream KLG balanced accuracy on OAI versus ImageNet MoCo pretraining size (log-scale x-axis). Performance shows a plateau: reducing pretraining from 1.2M to 6k yields minimal changes, indicating that sheer natural-image scale does not resolve the medical task. Error bars denote reported standard deviations.}
    \label{fig:imagenet_scaling}
\end{figure}

\begin{figure}[t]
    \centering
    \includegraphics[width=\columnwidth]{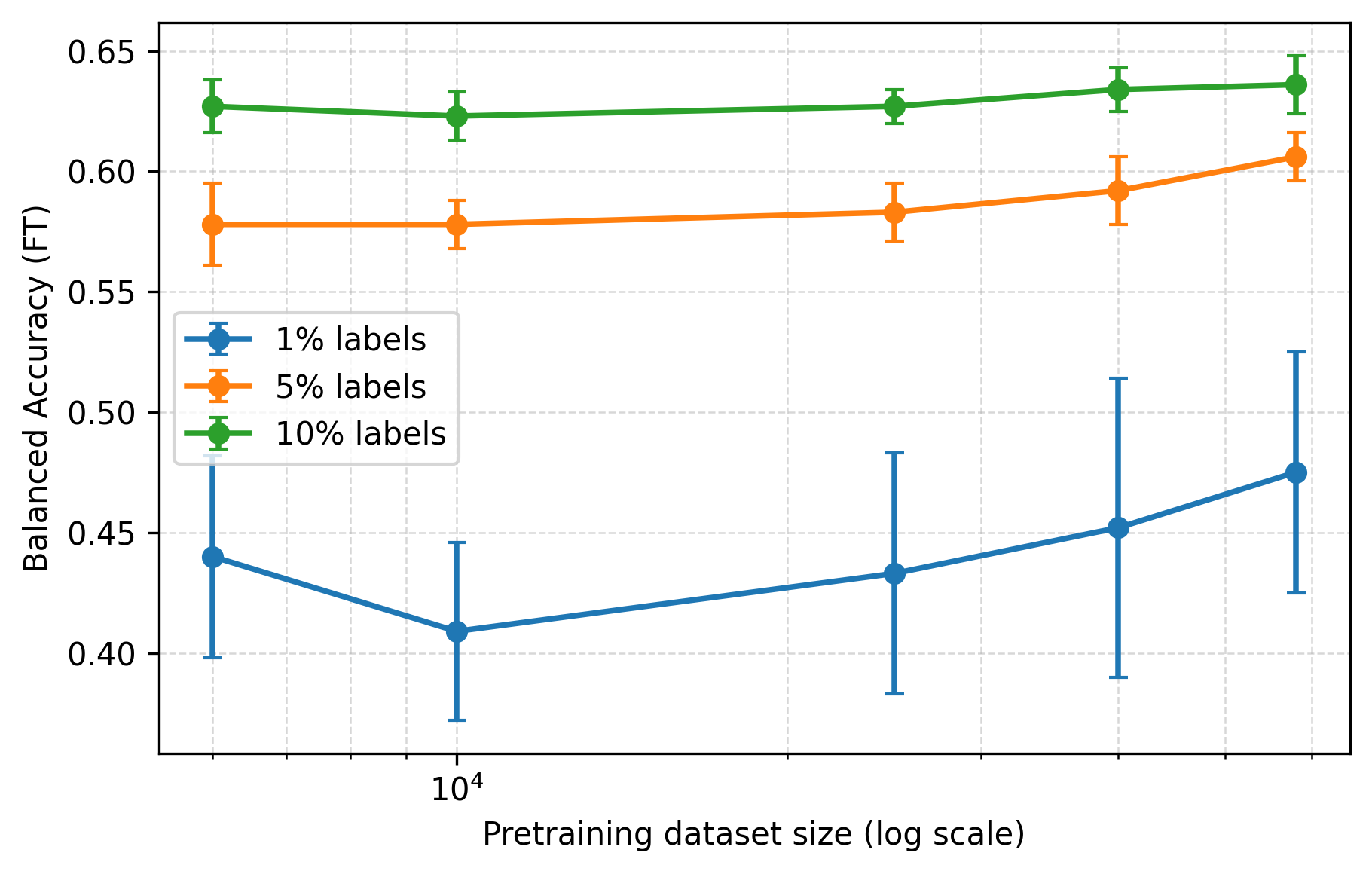}
    \caption{\textbf{In-domain scaling (Fine-tuning).} Downstream KLG balanced accuracy (FT) on OAI versus in-domain Knee MoCo pretraining size. Increasing in-domain pretraining data from 6k to 58k yields only marginal gains, consistent with a data-quality / signal bottleneck rather than data quantity. Error bars denote reported standard deviations.}
    \label{fig:knee_scaling_ft}
\end{figure}

\textbf{Result:} As shown in Figures \ref{fig:imagenet_scaling} and \ref{fig:knee_scaling_ft}, we observed a distinct performance plateau in both scenarios.
\begin{itemize}
    \item \textbf{ImageNet Scaling:} Reducing ImageNet size from Full (1.2M) to 58k resulted in negligible degradation (e.g., 0.648 vs. 0.627 at 10\% data), suggesting that vast scale is not required for the KLG task.
    \item \textbf{Knee Scaling:} More critically, increasing in-domain pretraining data from 6k to 58k also yielded minimal gains (e.g., 0.627 vs. 0.636 at 10\% data).
\end{itemize}
Together, these results confirm that the bottleneck is not the \textit{quantity} of pretraining data.

\subsubsection{The Task Complexity Hypothesis}
Having ruled out dataset size as the primary bottleneck, we investigated whether the limitation lay in the inherent \textit{complexity} of the downstream task itself.

\begin{figure}[t]
    \centering
    \includegraphics[width=\columnwidth]{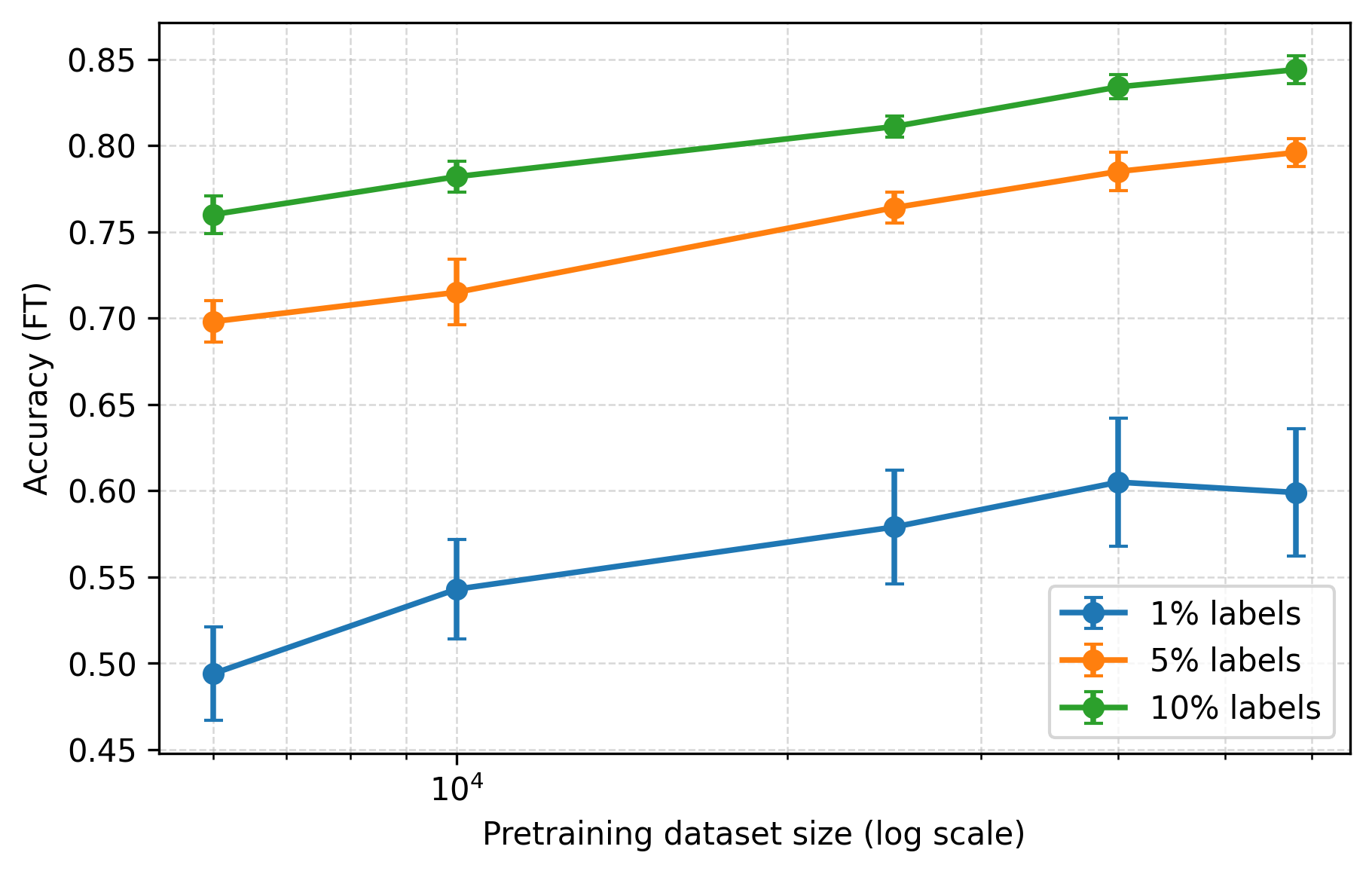}
    \caption{\textbf{Control Experiment: Imagenette Scaling.} Unlike the Knee domain, SSL (MoCo) shows strong, consistent scaling on simpler object-centric images (Imagenette). This confirms the difficulty lies in the medical task complexity, not the SSL method.}
    \label{fig:imagenette_scaling_ft}
\end{figure}

In stark contrast to the stagnant results on the Knee dataset, Figure \ref{fig:imagenette_scaling_ft} demonstrates strong, predictable scaling on Imagenette. The model achieves high accuracy (0.844) with just 10\% labeled data, and the performance improves consistently as the size of the pretraining data increases showing the expected scaling behavior.

\noindent\textbf{Additional controls (Appendix).} We further tested whether increasing visual diversity via mixed-domain medical pretraining (Knee+Chest) or cross-domain transfer could overcome the image-only bottleneck; these analyses did not yield consistent gains over ImageNet or in-domain knee pretraining (Appendix~\ref{appendix:A}). We also evaluated a generative image-only SSL approach (MAE) and observed similar diagnostic stagnation despite successful reconstructions (Appendix~\ref{appendix:B}).

\subsection{Cross-Domain Validation: Image-Only vs. Multimodal SSL}
To confirm that this limitation is not unique to Knee OA, we replicated experiments on Chest Radiography.

\textbf{Image-Only SSL Results:} Consistent with our Knee results, image-only SSL (MoCo, ViCReg) on Chest X-rays failed to outperform ImageNet initialization (Table \ref{tab:chest_image_only}).

\begin{table}[t]
\centering
\footnotesize
\caption{\textbf{Image-Only SSL Failure on Chest X-rays.} Macro AUROC on NIH Chest X-ray 14. Similar to Knee OA, pixel-only pretraining does not consistently beat ImageNet.}
\label{tab:chest_image_only}
\resizebox{\columnwidth}{!}{%
\begin{tabular}{lccc}
\toprule
\textbf{Pretraining} & \textbf{1\% (FT)} & \textbf{5\% (FT)} & \textbf{10\% (FT)} \\
\midrule
Imgnet-pret & 0.619 (0.013) & 0.699 (0.011) & 0.732 (0.009) \\
Chest MoCo & 0.572 (0.015) & 0.646 (0.012) & 0.680 (0.004) \\
Chest ViCReg & 0.558 (0.018) & 0.668 (0.006) & 0.699 (0.004) \\
\bottomrule
\end{tabular}%
}
\end{table}

\textbf{Multimodal SSL Results:} However, when we introduced text reports (Multimodal SSL) on the Chest dataset, we observed a significant performance jump. As shown in Table \ref{tab:chest_multimodal}, ConVIRT pretrained on CheXpert significantly outperformed ImageNet (e.g., 83.5 vs 72.6 LP at 1\% data). This validates that our multimodal implementation is correct and effective when the data distribution supports it.

\begin{table*}[t]
\centering
\footnotesize
\caption{\textbf{Multimodal SSL Success on Chest X-rays (Control Experiment).} When paired with reports, SSL (ConVIRT) significantly outperforms ImageNet (macro AUROC), proving the pipeline works in a standard setting.}
\label{tab:chest_multimodal}
\begin{tabular*}{\textwidth}{@{\extracolsep{\fill}}lcccccc}
\toprule
& \multicolumn{2}{c}{\textbf{1\% Data}} & \multicolumn{2}{c}{\textbf{5\% Data}} & \multicolumn{2}{c}{\textbf{10\% Data}} \\
\cmidrule(lr){2-3} \cmidrule(lr){4-5} \cmidrule(lr){6-7}
\textbf{Approach} & \textbf{LP} & \textbf{FT} & \textbf{LP} & \textbf{FT} & \textbf{LP} & \textbf{FT} \\
\midrule
Imgnet & 72.6 (0.8) & 75.9 (1.8) & 76.9 (0.9) & 82.8 (1.8) & 78.5 (0.7) & 83.8 (1.4) \\
ConVIRT & \textbf{83.5 (0.7)} & \textbf{79.5 (3.4)} & \textbf{85.1 (0.3)} & \textbf{85.7 (1.1)} & \textbf{85.5 (0.3)} & \textbf{86.2 (0.7)} \\
\bottomrule
\end{tabular*}
\end{table*}

\subsection{Knee Diagnosis Results and a Possible Explanation}

Despite the success in Chest X-rays, applying the same multimodal pipeline (ConVIRT, GLORIA) to the Knee dataset yielded weaker results for diagnosis (KL prediction).

\textbf{Failure at Diagnosis:} As shown in Table \ref{tab:knee_multimodal_diag}, multimodal pretraining substantially improves linear-probe performance over both ImageNet and image-only SSL (e.g., ConVIRT LP: 0.531 vs ImageNet LP: 0.320 at 10\% data), indicating that the pretrained representation captures disease-associated features present in the hospital cohort. However, these gains do not consistently translate to improved KL grading under fine-tuning (ConVIRT FT: 0.627 vs ImageNet FT: 0.646), suggesting a mismatch between the pretraining signal and the downstream diagnostic decision boundaries required for KLG prediction.

\textbf{Possible explanation: cohort construction.} A likely explanation is the way the NYU pretraining cohort was assembled. Because the hospital dataset was derived from adults with clinical knee OA diagnoses who underwent knee radiography or MRI in routine tertiary-care practice, it was likely enriched for symptomatic, treatment-seeking patients and may underrepresent the full disease spectrum needed for KL grading. In addition, radiology impressions describe findings but do not explicitly provide KL grades, which may further weaken alignment between multimodal pretraining and the downstream diagnosis task.

\begin{table*}[t]
\centering
\footnotesize
\caption{\textbf{Multimodal SSL Results for Knee Diagnosis.} Balanced Accuracy on OAI. Unlike Chest, Multimodal SSL fails to improve Diagnosis due to the lack of healthy controls in the pretraining data.}
\label{tab:knee_multimodal_diag}
\begin{tabular*}{\textwidth}{@{\extracolsep{\fill}}lcccccc}
\toprule
& \multicolumn{2}{c}{\textbf{1\% Data}} & \multicolumn{2}{c}{\textbf{5\% Data}} & \multicolumn{2}{c}{\textbf{10\% Data}} \\
\cmidrule(lr){2-3} \cmidrule(lr){4-5} \cmidrule(lr){6-7}
\textbf{Approach} & \textbf{FT} & \textbf{LP} & \textbf{FT} & \textbf{LP} & \textbf{FT} & \textbf{LP} \\
\midrule
Imgnet & 0.483 & 0.268 & 0.613 & 0.308 & \textbf{0.646} & 0.320 \\
ConVIRT & \textbf{0.570} & 0.455 & \textbf{0.619} & 0.520 & 0.627 & 0.531 \\
GLORIA & 0.400 & 0.411 & 0.558 & 0.463 & 0.589 & 0.482 \\
\bottomrule
\end{tabular*}
\end{table*}


\subsection{Knee OA Prognosis Results}
While the skewed distribution hindered diagnosis, we hypothesized it might be optimal for Prognosis, which requires differentiating between similarly diseased knees to predict progression.

\textbf{Superior Prognostic Performance:} Table \ref{tab:prognosis_results} presents performance on Structural Progression. Both multimodal approaches (ConVIRT, GLORIA) consistently improve AUROC over ImageNet and image-only SSL across label fractions, indicating that paired image–text pretraining is particularly well matched to prognostic discrimination among diseased knees. ConVIRT yields the strongest gains, approaching the \textbf{KL-Supervised Baseline} (pretrained on radiologist KL grades).

\begin{table*}[t]
\centering
\footnotesize
\setlength{\tabcolsep}{2pt} 
\caption{\textbf{Prognosis Results (Structural Progression and Incidence).} AUROC on OAI (Internal) and MOST (External). Multimodal SSL (ConVIRT) consistently outperforms ImageNet and Image-only SSL (MoCo, Barlow, ViCReg) across all data fractions, approaching the Supervised Upper Bound (KL Pret).}
\label{tab:prognosis_results}
\resizebox{\textwidth}{!}{%
\begin{tabular*}{\textwidth}{@{\extracolsep{\fill}}lcccccccccccc}
\toprule
\textbf{Approach} & \multicolumn{4}{c}{\textbf{5\% Data}} & \multicolumn{4}{c}{\textbf{10\% Data}} & \multicolumn{4}{c}{\textbf{25\% Data}} \\
\cmidrule(lr){2-5} \cmidrule(lr){6-9} \cmidrule(lr){10-13}
& \multicolumn{2}{c}{OAI} & \multicolumn{2}{c}{MOST} & \multicolumn{2}{c}{OAI} & \multicolumn{2}{c}{MOST} & \multicolumn{2}{c}{OAI} & \multicolumn{2}{c}{MOST} \\
& \textbf{LP} & \textbf{FT} & \textbf{LP} & \textbf{FT} & \textbf{LP} & \textbf{FT} & \textbf{LP} & \textbf{FT} & \textbf{LP} & \textbf{FT} & \textbf{LP} & \textbf{FT} \\
\midrule
Imgnet & 0.514 & 0.581 & 0.532 & 0.604 & 0.539 & 0.580 & 0.566 & 0.599 & 0.546 & 0.618 & 0.575 & 0.645 \\
MoCo & 0.572 & 0.577 & 0.583 & 0.601 & 0.590 & 0.603 & 0.612 & 0.636 & 0.620 & 0.633 & 0.661 & 0.679 \\
Barlow & 0.543 & 0.550 & 0.565 & 0.566 & 0.577 & 0.589 & 0.602 & 0.608 & 0.586 & 0.632 & 0.618 & 0.669 \\
ViCReg & 0.565 & 0.545 & 0.576 & 0.558 & 0.583 & 0.586 & 0.600 & 0.615 & 0.598 & 0.653 & 0.612 & 0.689 \\
CNN-JEPA &  0.497 & 0.586 & 0.519 & 0.620 & 0.574 & 0.640 & 0.611 & 0.670 & 0.573 & 0.622 & 0.599 & 0.648 \\

\textbf{ConVIRT} & \textbf{0.628} & \textbf{0.629} & \textbf{0.676} & \textbf{0.658} & \textbf{0.640} & \textbf{0.655} & \textbf{0.693} & \textbf{0.701} & \textbf{0.656} & \textbf{0.673} & \textbf{0.709} & \textbf{0.722} \\
GLORIA & 0.626 & 0.596 & 0.651 & 0.630 & 0.638 & 0.638 & 0.671 & 0.658 & 0.641 & 0.666 & 0.679 & 0.703 \\
\midrule
\textit{KL Pret (Sup)} & \textit{0.650} & \textit{0.679} & \textit{0.692} & \textit{0.724} & \textit{0.691} & \textit{0.694} & \textit{0.742} & \textit{0.745} & \textit{0.689} & \textit{0.716} & \textit{0.739} & \textit{0.766} \\
\bottomrule
\end{tabular*}%
}
\end{table*}

\section{Discussion}
In this study, we evaluated the utility of self-supervised learning (SSL) for knee osteoarthritis (OA) across diagnosis to longitudinal prognosis. Our results challenge the prevailing assumption in medical imaging that in-domain pretraining on unlabeled medical scans invariably yields superior representations compared to transfer learning from natural images (ImageNet). We further identify a critical dichotomy: while hospital radiographs and radiologist text impressions fail to improve diagnostic severity grading possibly due to selection bias, its effective for the prognosis task.

A primary finding of this work is the stagnation of image-only SSL (MoCo, Barlow Twins, ViCReg) for knee OA diagnosis. We observed a consistent discrepancy between Linear Probing (LP) and Fine-Tuning (FT) performance. In Linear Probing, in-domain SSL consistently outperformed ImageNet initialization. This confirms that in-domain pretraining successfully learned radiographic representations that are linearly separable and more semantically aligned with the downstream task than generic natural image features.

However, this advantage vanished under full Fine-Tuning, where ImageNet initialization matched or slightly exceeded the performance of in-domain SSL. Our control studies rule out pretraining dataset size as the bottleneck; rather, they point to task complexity. While SSL scales predictably on object-centric datasets like ImageNet (Figure \ref{fig:imagenette_scaling_ft}), it struggles to capture the subtle, continuous radiographic features of OA (e.g., joint space narrowing, osteophytes) without semantic guidance. This contrasts with prior studies in other domains, such as chest radiography \cite{sowrirajan2021moco,azizi2021big}, which notably reported substantial gains from in-domain pretraining.

We posit that this discrepancy is likely driven by differences in evaluation setup rather than domain characteristics alone. Unlike many previous works that utilize large, fixed validation sets even when training data is scarce, our protocol enforced proportional validation sampling and extensive Bayesian hyperparameter tuning for all methods, including the ImageNet baseline. Under these strictly controlled conditions, the performance gap narrows significantly. Our results suggest that standard ImageNet initialization is far more robust than typically credited in the medical SSL literature; when the supervised baseline is properly tuned, the marginal utility of learning pixel-level dependencies from in-domain datasets becomes negligible for classification tasks (both knee OA and chest tasks). Importantly, this negative result persists across additional image-only paradigms and data-diversity controls: mixed-domain medical pretraining (Knee+Chest) and masked autoencoder (MAE) pretraining do not consistently surpass ImageNet for KLG diagnosis (Appendix~\ref{appendix:A}--\ref{appendix:B}).

The introduction of radiology reports via multimodal SSL (ConVIRT, GLoRIA) highlighted the importance of alignment between the pretraining cohort and the downstream task. While we verified that ConVIRT improves chest X-ray classification (Table \ref{tab:chest_multimodal}), it did not improve knee KL grading, despite stronger linear-probe performance than ImageNet and image-only SSL methods. A plausible explanation is mismatch between the hospital pretraining corpus and the downstream diagnostic task. The NYU pretraining cohort was restricted to routine-care adults with clinically identified knee OA who underwent knee imaging at a tertiary care center, so the dataset did not cover the full spectrum from normal to severe disease required for balanced KL grading. In addition, radiology reports describe findings but do not explicitly encode KL grade. Together, these factors may have limited the ability of multimodal pretraining to improve fine-tuned KL grading.

At the same time, the same pretraining dataset was useful for prognosis. Structural incidence and progression require identifying patterns associated with future worsening, and hospital image-text pairs may provide useful signal about pathology variation relevant to this task. Consistent with this interpretation, ConVIRT pretrained on the hospital dataset outperformed ImageNet for structural progression and incidence prediction, and approached the performance of a supervised upper bound pretrained on expert KL grade labels.

\begin{figure*}[t!]
    \centering
    \begin{subfigure}[b]{0.485\textwidth}
        \centering
        \includegraphics[width=\linewidth]{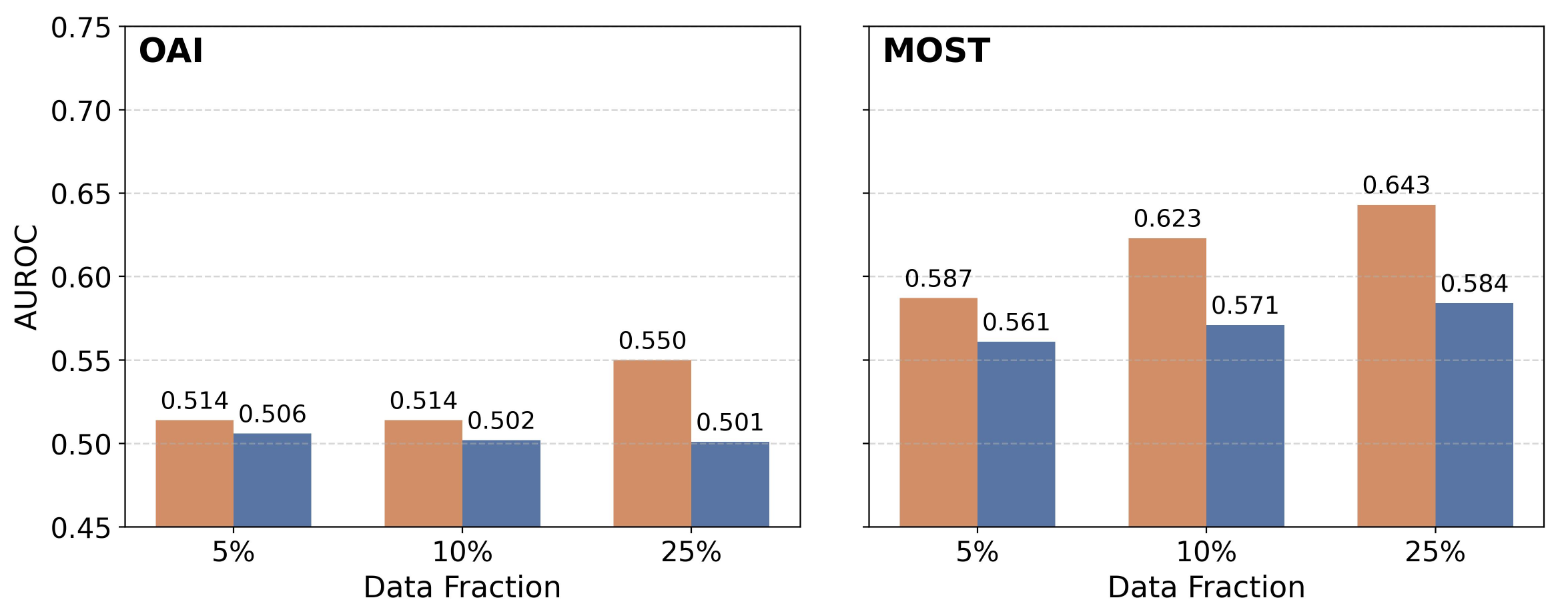}
        \caption{KL 0}
        \label{fig:kl0}
    \end{subfigure}
    \hfill
    \begin{subfigure}[b]{0.485\textwidth}
        \centering
        \includegraphics[width=\linewidth]{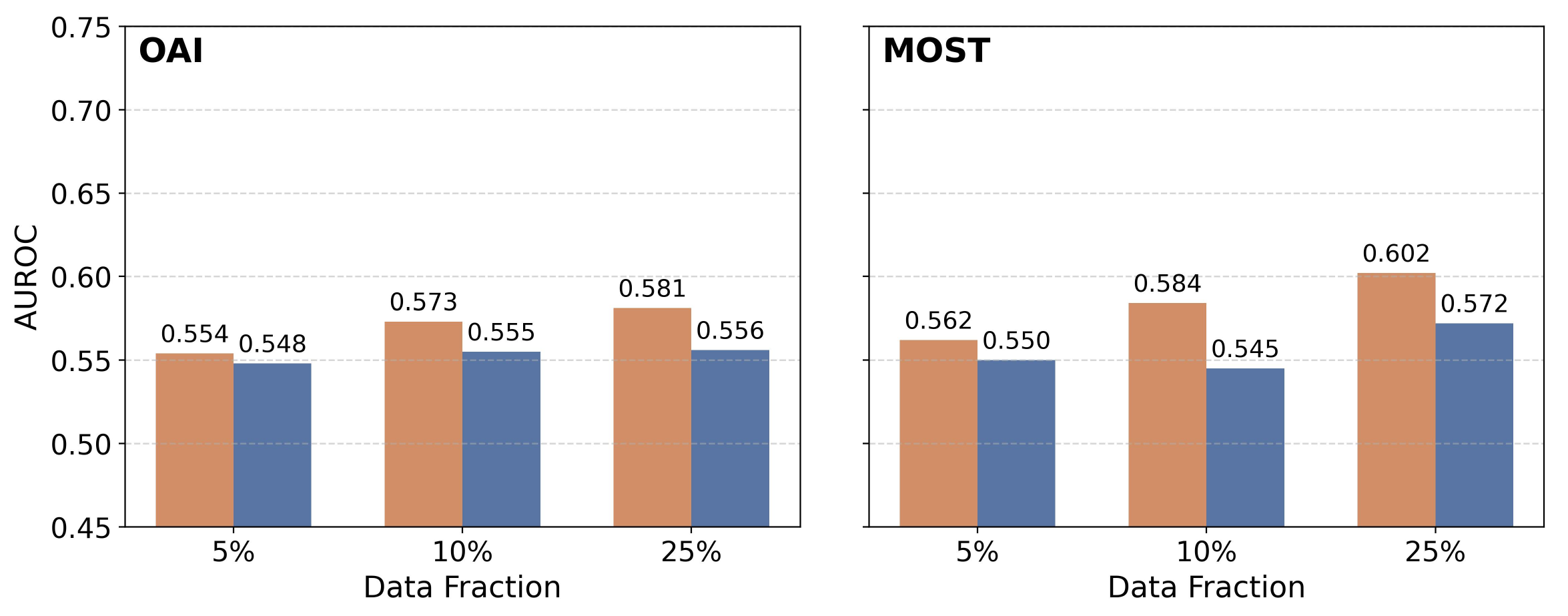}
        \caption{KL 1}
        \label{fig:kl1}
    \end{subfigure}

    \vspace{0.9em}

    \begin{subfigure}[b]{0.485\textwidth}
        \centering
        \includegraphics[width=\linewidth]{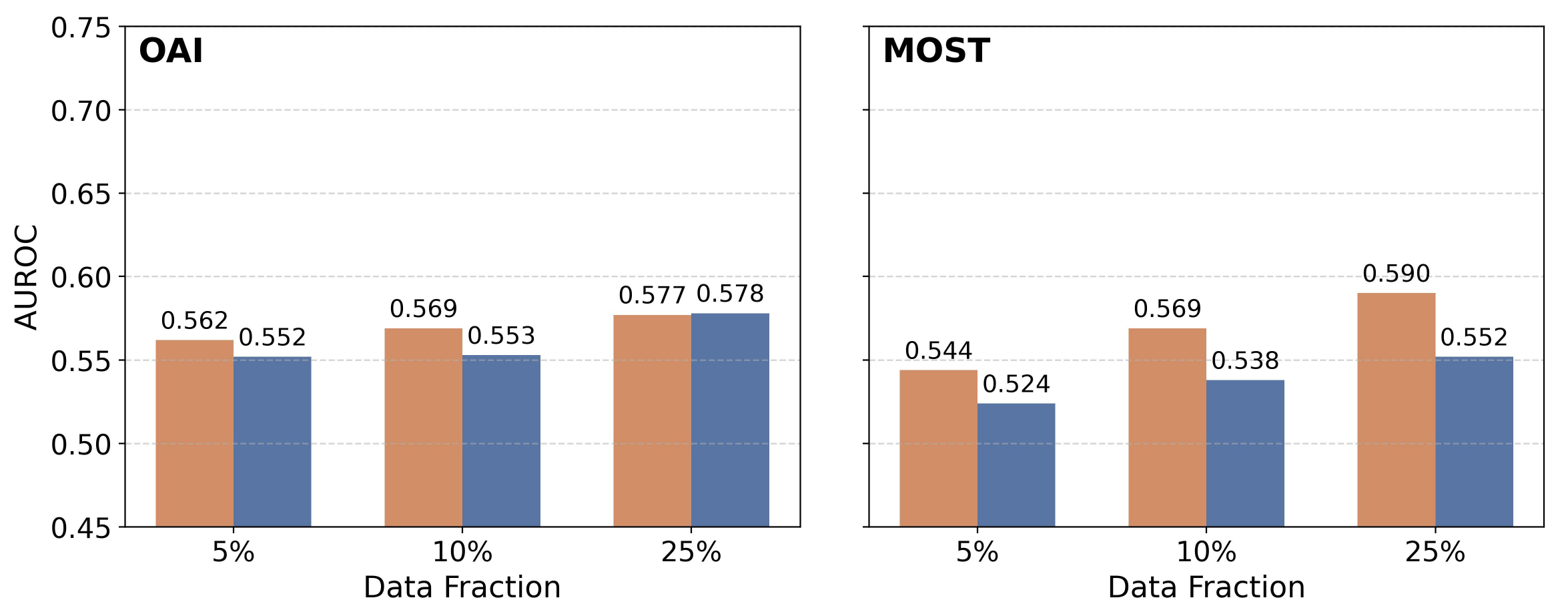}
        \caption{KL 2}
        \label{fig:kl2}
    \end{subfigure}
    \hfill
    \begin{subfigure}[b]{0.485\textwidth}
        \centering
        \includegraphics[width=\linewidth]{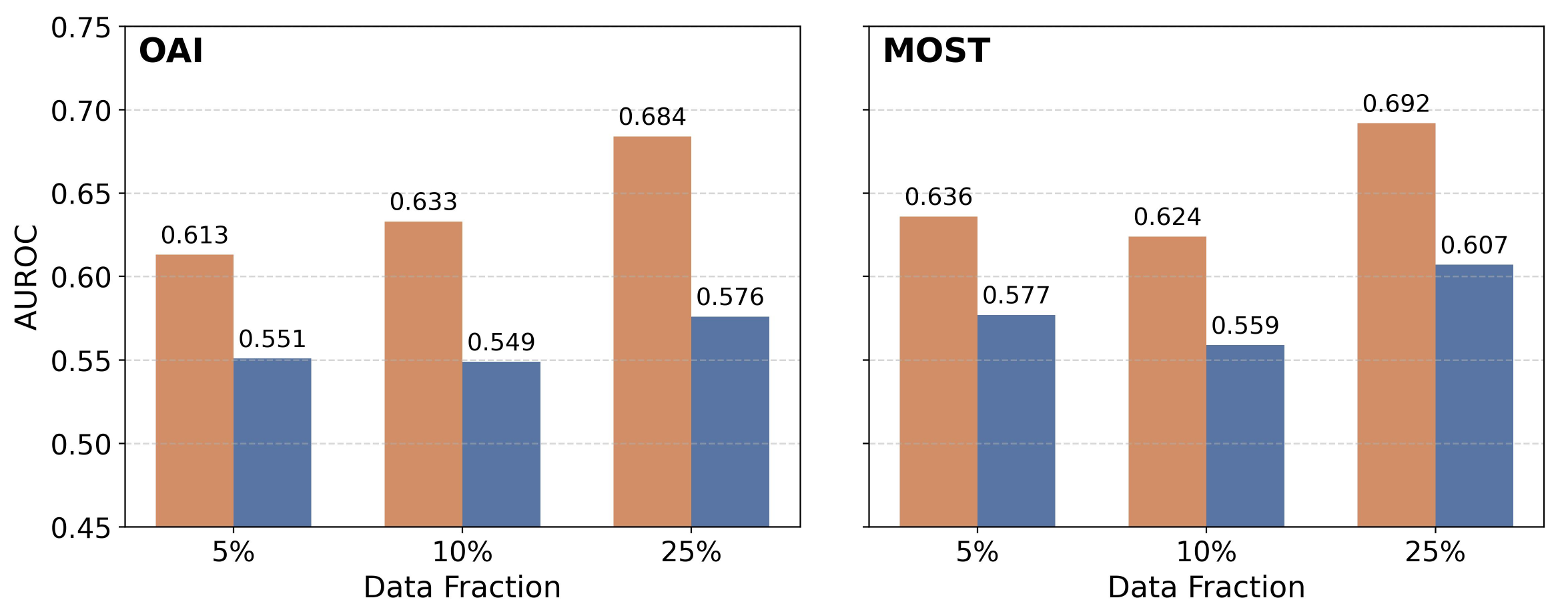}
        \caption{KL 3}
        \label{fig:kl3}
    \end{subfigure}

    \vspace{0.35em}
    \centering
    \includegraphics[height=0.8cm]{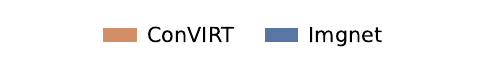}

    \caption{\textbf{Stratified prognostic performance by baseline severity (ImageNet vs. ConVIRT).} Performance (AUROC) on the structural progression/incidence task stratified by baseline Kellgren-Lawrence (KL) grade in the external validation set (MOST). Multimodal pretraining (ConVIRT) consistently outperforms ImageNet across baseline severity strata, suggesting that the learned representation captures features relevant to future structural worsening across disease stages.}
    \label{fig:stratified_prognosis}
\end{figure*}

As shown in Figure \ref{fig:stratified_prognosis}, multimodal pretraining improved discrimination across baseline severity levels, including both early-stage disease (KL 0/1) and established OA (KL 2/3). In the external validation cohort (MOST) at 10\% labeled data, ConVIRT outperformed ImageNet in the KL 0 and KL 1 subgroups as well as in more advanced strata. Although the hospital pretraining cohort was limited to OA knees, the gains across strata suggest that the learned representation captures structural patterns that are broadly relevant to prognosis rather than being confined to a single severity subgroup.

These findings suggest that the utility of routine-care hospital data is task-dependent. For diagnostic tasks such as KL grading, a pretraining cohort restricted to clinically identified OA knees may provide limited coverage of the full disease spectrum needed for balanced classification. For prognostic tasks, however, the same data may still be valuable when the downstream objective is more closely aligned with the pathology patterns represented in routine clinical care.

Second, our results demonstrate a scalable path for developing prognostic models. Validated prognostic datasets like OAI are small and expensive to curate. By leveraging readily available image-text pairs from hospital archives, we achieved performance competitive with fully supervised pretraining without requiring manual annotation. This approach could significantly reduce the cost of developing prognostic models for clinical trials and treatment selection.

Our study has several limitations. First, while the pretraining data (NYU) was large, it comes from a single medical center, potentially limiting geographic generalizability. Second, the OAI and MOST cohorts, which are derived from standardized research protocols, potentially lack the variability of routine clinical practice and therefore affecting generalizability to diverse care settings. They also lack racial diversity. As shown in Table \ref{tab:progression_demographics}, the OAI progression cohort is predominantly White (184 men, 253 women) compared to Black participants (26 men, 90 women). Similarly, the incidence cohort (Table \ref{tab:incidence_demographics}) contains only 13 Black men compared to 133 White men. This imbalance may mask performance disparities across racial groups, and future work must validate these models on more diverse populations to ensure equitable performance. Finally, we relied on 2D radiographs; while MRI provides higher sensitivity for early OA, radiographs remain the standard of care for initial assessment, ensuring the clinical relevance of our findings.

We conclude that while image-only self-supervision has limits in medical imaging, multimodal pretraining offers a powerful mechanism for learning prognostic representations. By aligning the pretraining data distribution with the downstream clinical goal and leveraging the abundance of pathological examples in hospital records, we can build robust prognostic models that transcend the limitations of small, curated datasets.


\section{Conclusions}

In summary, we found that image-only self-supervised learning did not consistently improve knee osteoarthritis diagnosis over ImageNet pretraining under full fine-tuning, despite gains in linear probing. In contrast, multimodal pretraining on hospital radiographs paired with radiology impressions improved prognostic modeling, including external validation on MOST. These findings suggest that the usefulness of large-scale hospital data depends on the downstream task. In our setting, routine-care image-text data drawn from patients with clinically identified OA appeared less well matched to KL grading, which requires coverage of the full disease spectrum, but remained informative for prognosis, where the downstream objective was better aligned with the pretraining cohort. More broadly, this work points to a scalable approach for developing clinically relevant prognostic models from routinely collected data without extensive manual annotation.

\bibliographystyle{model1-num-names}
\bibliography{cas-refs}
\appendix

\renewcommand{\thefigure}{\thesection.\arabic{figure}}
\renewcommand{\thetable}{\thesection.\arabic{table}}
\setcounter{figure}{0} 
\setcounter{table}{0}
\section{Appendix - Extended Pretraining Analysis: Data Diversity and Domain Transfer}
\label{appendix:A}

Building upon the main findings, we conducted further investigations to determine if the lack of improvement from image-only SSL was due to limited visual diversity in the knee radiograph dataset. We hypothesized that combining diverse medical domains (Knee + Chest) might learn more robust general-purpose features than knee data alone.

\subsection{Methods Explored}
We compared several additional SSL configurations:
\begin{itemize}
    \item \textbf{Pretraining Data Variants:}
    \begin{itemize}
        \item \textbf{Knee Domain Mix:} The standard pretraining set (~58k images) from OAI, MOST, and NYU.
        \item \textbf{Chest Domain:} Pretraining solely on NIH Chest X-ray 14.
        \item \textbf{Knee + Chest:} A combined pretraining dataset to test if visual diversity improves downstream transfer.
    \end{itemize}
    \item \textbf{Evaluation:} We evaluated transfer performance on both the Knee KLG task (OAI) and the Chest pathology task (NIH/CheXpert) to assess cross-domain transferability.
\end{itemize}

\subsection{Key Findings}
As detailed in Tables \ref{tab:appendix_klg_diversity} and \ref{tab:appendix_chest_diversity}, the results indicate that increasing visual diversity via domain mixing did not unlock performance gains:
\begin{itemize}
    \item \textbf{Diversity did not replace Semantics:} The \textit{Knee+Chest} models did not significantly outperform pure \textit{Knee} models or \textit{ImageNet} baselines on KLG grading (Table \ref{tab:appendix_klg_diversity}).
    \item \textbf{Poor Cross-Domain Transfer:} Models pretrained on Chest X-rays performed poorly on Knee tasks, and vice-versa (Table \ref{tab:appendix_chest_diversity}), confirming that pixel-level features are highly domain-specific and do not generalize well without semantic alignment.
    \item \textbf{ImageNet Baseline Strength:} ImageNet pretraining consistently rivaled or beat even the "diverse" medical pretraining (Knee+Chest), reinforcing that natural image features are a sufficiently robust starting point for medical tasks when fine-tuning is employed.
\end{itemize}

\begin{table*}[t]
\centering
\footnotesize
\caption{\textbf{Extended KLG Diagnosis Results (OAI Test Set).} Comparison of pretraining sources, including cross-domain (Chest) and mixed-domain (Knee+Chest) initialization. Adding Chest data to increase diversity (Knee+Chest) did not yield significant gains over ImageNet.}
\label{tab:appendix_klg_diversity}
\begin{tabular*}{\textwidth}{@{\extracolsep{\fill}}lcccccc}
\toprule
& \multicolumn{2}{c}{\textbf{1\% Data}} & \multicolumn{2}{c}{\textbf{5\% Data}} & \multicolumn{2}{c}{\textbf{10\% Data}} \\
\cmidrule(lr){2-3} \cmidrule(lr){4-5} \cmidrule(lr){6-7}
\textbf{Pretraining Approach} & \textbf{FT} & \textbf{LP} & \textbf{FT} & \textbf{LP} & \textbf{FT} & \textbf{LP} \\
\midrule
Random & 0.397 (0.042) & 0.200 & 0.555 (0.011) & 0.200 & 0.619 (0.015) & 0.200 \\
Imgnet-pret & 0.483 (0.040) & 0.263 & 0.613 (0.014) & 0.308 & 0.646 (0.009) & 0.320 \\
\midrule
\textit{In-Domain (Knee)} & & & & & & \\
MoCo (Knee) & 0.475 (0.050) & 0.368 & 0.606 (0.010) & 0.411 & 0.636 (0.012) & 0.422 \\
Barlow (Knee) & 0.425 (0.063) & 0.324 & 0.589 (0.018) & 0.355 & 0.620 (0.010) & 0.358 \\
ViCReg (Knee) & 0.471 (0.048) & 0.336 & 0.585 (0.018) & 0.374 & 0.635 (0.007) & 0.373 \\
\midrule
\textit{Cross-Domain (Chest)} & & & & & & \\
Chest MoCo & 0.283 (0.064) & - & 0.563 (0.031) & - & 0.611 (0.009) & - \\
Chest ViCReg & 0.330 (0.050) & - & 0.575 (0.010) & - & 0.623 (0.007) & - \\
\midrule
\textit{Mixed-Domain (Knee+Chest)} & & & & & & \\
Knee+Chest MoCo & 0.363 (0.033) & - & 0.577 (0.010) & - & 0.636 (0.013) & - \\
Knee+Chest ViCReg & 0.330 (0.050) & - & 0.590 (0.014) & - & 0.613 (0.009) & - \\
\bottomrule
\end{tabular*}
\end{table*}

\begin{table*}[t]
\centering
\footnotesize
\caption{\textbf{Extended Chest Pathology Results (NIH Test Set).} Cross-domain evaluation showing that Knee pretraining fails to transfer effectively to Chest X-ray tasks compared to ImageNet.}
\label{tab:appendix_chest_diversity}
\begin{tabular*}{\textwidth}{@{\extracolsep{\fill}}lcccccc}
\toprule
& \multicolumn{2}{c}{\textbf{1\% Data}} & \multicolumn{2}{c}{\textbf{5\% Data}} & \multicolumn{2}{c}{\textbf{10\% Data}} \\
\cmidrule(lr){2-3} \cmidrule(lr){4-5} \cmidrule(lr){6-7}
\textbf{Pretraining Approach} & \textbf{FT} & \textbf{LP} & \textbf{FT} & \textbf{LP} & \textbf{FT} & \textbf{LP} \\
\midrule
Imgnet-pret & 0.619 (0.013) & 0.583 & 0.699 (0.011) & 0.614 & 0.732 (0.009) & 0.633 \\
\midrule
\textit{In-Domain (Chest)} & & & & & & \\
Chest MoCo & 0.572 (0.015) & 0.593 & 0.646 (0.012) & 0.632 & 0.680 (0.004) & 0.648 \\
Chest Barlow & 0.601 (0.018) & 0.570 & 0.676 (0.006) & 0.598 & 0.702 (0.006) & 0.615 \\
\midrule
\textit{Cross-Domain (Knee)} & & & & & & \\
Knee MoCo & 0.544 (0.015) & - & 0.661 (0.010) & - & 0.684 (0.010) & - \\
Knee ViCReg & 0.551 (0.017) & - & 0.654 (0.007) & - & 0.668 (0.008) & - \\
\midrule
\textit{Mixed-Domain (Knee+Chest)} & & & & & & \\
Knee+Chest MoCo & 0.564 (0.008) & - & 0.613 (0.010) & - & 0.662 (0.008) & - \\
Knee+Chest Barlow & 0.598 (0.010) & - & 0.676 (0.007) & - & 0.696 (0.010) & - \\
\bottomrule
\end{tabular*}
\end{table*}


\setcounter{figure}{0} 
\setcounter{table}{0}
\section{Appendix - Additional Image-Only SSL: Masked Autoencoders (MAE)}
\label{appendix:B}

To further explore the potential of image-only self-supervised learning, we evaluated Masked Autoencoders (MAE), a generative approach that learns representations by reconstructing randomly masked patches of an image \cite{he2022masked}. In these experiments, we use a \textbf{ViT-Small} encoder (following the standard MAE formulation) rather than the ResNet-34 backbone used in the main knee experiments; results are therefore reported as an additional control within this MAE-specific setup. Unlike contrastive methods (MoCo, Barlow Twins) which focus on global discrimination, MAE forces the model to learn local anatomical context.

\subsection{Experimental Setup}
We pretrained MAE models on the combined knee radiograph dataset, varying the masking ratio from 0.25 to 0.7. We evaluated these encoders on the KLG diagnosis task (OAI) and qualitatively assessed their reconstruction capabilities.

\subsection{Results: The Reconstruction Paradox}
The quantitative results (Table \ref{tab:appendix_mae}) mirror our previous findings: MAE pretraining did not consistently outperform the ImageNet baseline. For instance, at 10\% labeled data, the best MAE model (masking ratio 0.5) achieved a balanced accuracy of 0.609, slightly underperforming ImageNet (0.618).

\begin{figure*}[t]
    \centering
    \includegraphics[width=0.8\textwidth]{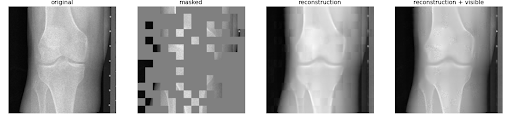}
    \caption{\textbf{The Reconstruction Paradox (MAE Visualizations).} Sample reconstructions with a masking rate of 0.6, showing the original input, masked view, model reconstruction, and overlay. The model demonstrates high fidelity in hallucinating anatomical structures from partial data, yet this structural understanding did not translate to improved KLG diagnostic performance.}
    \label{fig:mae_viz}
\end{figure*}

However, the qualitative analysis reveals a compelling paradox. As shown in Figure \ref{fig:mae_viz}, the MAE models successfully reconstructed high-fidelity anatomical details, including joint space and bone contours, even when 60\% of the image was masked. 

\begin{itemize}
    \item \textbf{Anatomical Prior vs. Pathological Feature:} The high quality of reconstruction indicates the model learned a strong statistical prior of "healthy" knee anatomy. It can effectively "fill in" the missing pixels based on the general structure of a knee.
    \item \textbf{Diagnostic Failure:} Despite this understanding of structure, the downstream failure suggests that the features used for reconstruction (macroscopic geometry) are distinct from those required for diagnosis (subtle osteophytes, specific joint space narrowing patterns). The model learned to be an anatomical artist, but not a diagnostician.
\end{itemize}

    
\begin{table*}[t]
\centering
\footnotesize
\caption{\textbf{Masked Autoencoder (MAE) Pretraining Results.} Comparison of ImageNet pretraining vs. MAE pretraining with varying masking rates. Performance is reported as Balanced Accuracy (mean (std)) on the OAI test set. Despite successful image reconstruction, diagnostic performance remains stagnant.}
\label{tab:appendix_mae}
\begin{tabular*}{\textwidth}{@{\extracolsep{\fill}}lcccccc}
\toprule
& \multicolumn{2}{c}{\textbf{1\% Data}} & \multicolumn{2}{c}{\textbf{5\% Data}} & \multicolumn{2}{c}{\textbf{10\% Data}} \\
\cmidrule(lr){2-3} \cmidrule(lr){4-5} \cmidrule(lr){6-7}
\textbf{Pretraining Approach} & \textbf{FT} & \textbf{LP} & \textbf{FT} & \textbf{LP} & \textbf{FT} & \textbf{LP} \\
\midrule
Imgnet-pret & 0.489 (0.052) & 0.301 (0.017) & 0.589 (0.013) & 0.345 (0.013) & 0.618 (0.013) & 0.381 (0.010) \\
\midrule
MAE 0.25 & 0.398 (0.098) & 0.247 (0.011) & 0.574 (0.012) & 0.305 (0.005) & 0.603 (0.011) & 0.332 (0.005) \\
MAE 0.35 & 0.498 (0.024) & 0.260 (0.011) & 0.571 (0.012) & 0.324 (0.006) & 0.581 (0.021) & 0.352 (0.005) \\
MAE 0.5 & 0.399 (0.095) & 0.287 (0.013) & 0.573 (0.007) & 0.358 (0.012) & 0.609 (0.009) & 0.389 (0.012) \\
MAE 0.6 & 0.464 (0.066) & 0.295 (0.012) & 0.581 (0.017) & 0.373 (0.015) & 0.594 (0.012) & 0.404 (0.017) \\
MAE 0.7 & 0.445 (0.050) & 0.299 (0.015) & 0.553 (0.011) & 0.374 (0.015) & 0.590 (0.011) & 0.405 (0.015) \\
\bottomrule
\end{tabular*}
\end{table*}

\section{Appendix - Implementation Details and Hyperparameters}
\label{appendix:C}

To ensure reproducibility, we detail the specific architectures, initialization strategies, and the automated hyperparameter tuning protocol used in our experiments.

\subsection{Model Architectures}
We selected backbone architectures consistent with standard benchmarks in each respective domain:
\begin{itemize}
    \item \textbf{Knee OA Experiments:} A \textbf{ResNet-34} backbone was used for all knee experiments (both image-only and multimodal) to match prior baselines in osteoarthritis grading.
    \item \textbf{Chest Radiography Experiments:} A \textbf{ResNet-50} backbone was used for all chest X-ray control experiments, consistent with standard evaluation protocols on CheXpert and NIH Chest X-ray 14.
\end{itemize}

\subsection{Pretraining Configuration}

\subsubsection{Image-Only SSL (MoCo, Barlow Twins, ViCReg)}
We utilized the official reference implementations for each self-supervised method. 
\begin{itemize}
    \item \textbf{Training Schedule:} All image-only models were pretrained for 100 epochs.
    \item \textbf{Optimization:} We utilized the default hyperparameter configurations provided in the official repositories, with the learning rate tuned specifically to achieve convergence on our datasets.
\end{itemize}

\subsubsection{Multimodal SSL (ConVIRT, GLoRIA)}
For image-text pretraining, we employed a transfer learning initialization strategy to accelerate convergence:
\begin{itemize}
    \item \textbf{Image Encoder:} Initialized with standard ImageNet pretrained weights.
    \item \textbf{Text Encoder:} Initialized with \textbf{Bio\_ClinicalBERT} \cite{alsentzer2019publicly} weights to leverage domain-specific medical language understanding.
\end{itemize}

\subsection{Data Augmentation Pipeline}
We utilized a consistent set of data augmentations during pretraining to encourage invariance to standard radiographic variations. The pipeline consisted of:
\begin{enumerate}
    \item \textbf{Random Resized Crop:} Crops were resized to $224 \times 224$ pixels with a scale range of $(0.5, 1.0)$.
    \item \textbf{Gaussian Blur:} Applied with a probability of $p=0.1$ and a radius $\sigma \in [0.1, 2.0]$.
    \item \textbf{Random Horizontal Flip:} Applied with a probability of $p=0.5$.
    \item \textbf{Random Rotation:} Images were rotated by a random angle in the range $[-25^\circ, +25^\circ]$.
\end{enumerate}

During finetuning Random Resized Crop, Random Horizontal Flip and Random Rotation were used. For evaluation, the input was resized to  $256 \times 256$ and center cropped to $224 \times 224$.

\subsection{Downstream Fine-Tuning Protocol}
For all downstream tasks, we utilized an automated Bayesian hyperparameter search rather than fixed settings. This was implemented using the \textbf{Optuna} framework with the following specific configuration:

\subsubsection{Hyperparameter Search Space}
For every unique combination of pretraining method (e.g., MoCo, ImageNet) and data fraction (1\%, 5\%, 10\%), we ran 50 trials using a Tree-structured Parzen Estimator (TPE) sampler. The search space was defined as follows:
\begin{itemize}
    \item \textbf{Optimizer:} Categorical selection between \texttt{Adam} \cite{kingma2014adam}, \texttt{SGD}, and \texttt{AdamW} \cite{loshchilov2017decoupled}.
    \item \textbf{Learning Rate (LR):} Log-uniform distribution in the range $[1e^{-6}, 1e^{-4}]$.
    \item \textbf{Weight Decay:} Log-uniform distribution in the range $[1e^{-6}, 1e^{-3}]$.
    \item \textbf{Training Epochs:} Integer range $[40, 120]$.
\end{itemize}

\subsubsection{Learning Rate Scheduling}
We utilized a \textbf{One-Cycle Learning Rate Policy} for all fine-tuning runs to accelerate convergence. The maximum learning rate for the cycle was dynamically coupled to the base LR via a tunable multiplier:
\begin{equation}
    LR_{max} = LR_{base} \times T_{mult}
\end{equation}
where $T_{mult}$ was a hyperparameter optimized in the integer range $[1, 5]$.

\subsubsection{Pruning Strategy}
To improve search efficiency during Bayesian hyperparameter tuning, we employed a \textbf{Median Pruner}. Trials were terminated early if their validation accuracy at any epoch fell below the median of previous trials at the same step. A warmup period of 5 epochs was allowed before pruning began.
\end{document}